\documentclass{article}

\usepackage{arxiv}
\usepackage[utf8]{inputenc} 
\usepackage[T1]{fontenc}    
\usepackage{hyperref}       
\usepackage{url}            
\usepackage{booktabs}       
\usepackage{amsfonts}       
\usepackage{nicefrac}       
\usepackage{microtype}      
\usepackage{lipsum}
\usepackage{graphicx}
\graphicspath{ {./images/} }
\usepackage{subcaption}
\usepackage{adjustbox}
\usepackage{amsmath}
\usepackage{multirow}
\usepackage{makecell}

\begin{document}

\title{UMM-RM: An Upcycle-and-Merge MoE Reward Model for Mitigating Reward Hacking}
\author{
Lingling Fu$^{1}$ \quad Yongfu Xue$^{2}$\\
$^{1}$Guangxi University\\
\texttt{2313593009@st.gxu.edu.cn, xueyongfu@outlook.com}
}
	
\maketitle
	
\begin{abstract}
Reward models (RMs) are critical to reinforcement learning from human feedback (RLHF). However, policy models can exploit biases or spurious correlations in dense RMs, causing reward hacking: RM scores rise while alignment with human preferences deteriorates, especially under distribution shift. We propose UMM-RM (Upcycle-and-Merge MoE Reward Model), which copies the feed-forward layers of an SFT checkpoint to initialize shared and routed experts, then trains the model on preference data. The always-active shared expert captures instruction-agnostic preference signals, while routed experts model fine-grained preferences across instructions or tasks. The experts are then merged into a single dense RM using learnable weights, preserving the robustness and exploitation resistance of expert diversity without MoE or ensemble inference overhead. Across multiple base models and preference datasets, UMM-RM improves preference accuracy, reduces reward hacking during PPO training, and achieves more stable preference alignment than standard dense RMs.
\end{abstract}
	

\section{Introduction}

Large language models have made remarkable progress in question answering, reasoning, and code generation~\cite{brown2020language,touvron2023llama}, yet pretraining alone does not ensure alignment with human preferences and values, leaving models prone to unsafe, biased, or misaligned outputs~\cite{ouyang2022training}. Preference alignment typically combines supervised fine-tuning (SFT) with offline preference optimization, such as DPO~\cite{rafailov2023direct}, or online RLHF. We focus on PPO-based RLHF~\cite{Proximal2017Schulman}, where a learned reward model guides policy optimization.

Despite the success of RLHF, reward models remain vulnerable to reward hacking. When the learned reward imperfectly reflects human preferences, the policy may exploit shortcut features or model biases to increase reward without improving true response quality~\cite{Scaling2023Gao}. As optimization proceeds, such errors can be amplified, causing proxy rewards to diverge from actual preference quality.

Existing methods mitigate this problem through more robust reward-model training~\cite{yang2024regularizing,jia2024generalizing} or uncertainty-aware and ensemble-based optimization~\cite{Reward2023Coste,Scalable2024Ahmed}. Ensembles that aggregate the outputs of multiple reward models incur additional inference overhead, whereas WARM~\cite{Warm2024Rame} reduces this overhead through parameter averaging. These approaches can nevertheless remain vulnerable when individual reward models share similar biases.

To address these limitations, we propose UMM-RM, an Upcycle-and-Merge Mixture-of-Experts reward model that adapts the upcycle-and-merge paradigm of XFT~\cite{XFT2024Ding} from code instruction tuning to preference-based reward modeling. UMM-RM starts from an SFT checkpoint, copies its feed-forward layers to initialize an MoE architecture with an always-active shared expert and routed experts, and trains the resulting model on preference data. During MoE training, the shared expert has a fixed gating weight and is intended to capture general preference signals, while routed experts model fine-grained preferences across instructions and tasks. The trained experts are then merged into a dense reward model through preference-calibrated parameter aggregation, aiming to reduce opportunities for the policy to exploit individual experts' biases and thereby mitigate reward hacking during PPO while retaining dense-model inference efficiency.

Our contributions are as follows:
\begin{itemize}
    \item We propose an upcycle-and-merge reward modeling framework combining a shared-preference expert with preference-calibrated merging to mitigate reward hacking while retaining dense-model inference efficiency.
    
    \item We evaluate UMM-RM across multiple base models and expert configurations, demonstrating its effectiveness in mitigating reward hacking during PPO.
    
    \item UMM-RM consistently improves accuracy across preference datasets and achieves competitive end-to-end alignment against ensemble and MoE reward models at dense-level inference cost.
\end{itemize}
	
\section{Background}

\subsection{Reward Modeling}

Reward models learn human preferences from pairwise preference data. Given a prompt \(x\), a preferred response \(y_w\), and a rejected response \(y_l\), a standard reward model is trained using the Bradley--Terry objective~\cite{Rank1952Bradley}:

\begin{equation}
\begin{aligned}
\mathcal{L}(R)
= -\mathbb{E}_{(x, y_w, y_l) \in D}
\Big[
&\log \sigma\big(
R(x, y_w) 
&- R(x, y_l)
\big)
\Big].
\end{aligned}
\end{equation}
This objective encourages the model to assign higher reward scores to preferred responses than to rejected ones.

\subsection{PPO in RLHF}

Proximal Policy Optimization (PPO)~\cite{Proximal2017Schulman} updates the policy to maximize rewards from the learned reward model. To limit divergence from the reference SFT policy, we use the following KL-regularized reward:
\begin{equation}
R^{\text{PPO}}(q,a)
=R(q,a)
-\beta \log
\frac{\pi^{\text{PPO}}(a\mid q)}
{\pi^{\text{init}}(a\mid q)},
\end{equation}
where \(R(q,a)\) is the learned reward, \(\pi^{\text{PPO}}\) is the policy being optimized, \(\pi^{\text{init}}\) is the reference SFT policy, and \(\beta\) controls the strength of KL regularization.

\subsection{Mixture-of-Experts}

Mixture-of-Experts (MoE) architectures replace a single feed-forward network (FFN) with multiple experts and use a router to sparsely activate a subset of them. This increases model capacity with sublinear compute growth~\cite{Outrageously2017Shazeer} and can improve model capability~\cite{Moefication2021Zhang}. An MoE layer is expressed as
\begin{equation}
y=\sum_{i=1}^{n}G(x)_i E_i(x),
\end{equation}
where \(G(x)_i\) denotes the routing weight and \(E_i(x)\) the output of expert \(i\).

\subsection{Reward Hacking}

In RLHF, the reward model only approximates human preferences, allowing the policy to exploit inaccuracies in the learned reward signal. As optimization proceeds, the proxy reward may continue increasing while true alignment deteriorates, indicating reward hacking. A common diagnostic is to evaluate the same policy outputs with a stronger reference reward model (Gold RM): divergence between rising proxy rewards and declining gold rewards signals over-optimization. Although scaling the proxy reward model and increasing training data can reduce this effect~\cite{Scaling2023Gao}, such approaches substantially increase computational cost.

\section{Method}
\subsection{Motivation}

Reward hacking can arise when a reward model overfits local patterns or spurious features, assigning high rewards to outputs that do not truly align with human preferences~\cite{ouyang2022training,Scaling2023Gao}. This issue can become more pronounced in MoE reward models, where individual experts specialize on different regions of the input space.

Although expert specialization increases modeling capacity, some experts may rely on superficial patterns such as output length, templated expressions, or other task-specific correlations. When the router repeatedly selects such experts for particular inputs, their anomalous reward responses can be exploited by the policy during PPO, amplifying reward bias and destabilizing optimization.

This risk is further increased by the relatively limited scale of reward-model data, which leaves individual experts with fewer effective training samples and higher estimator variance. We therefore introduce an always-active shared expert to capture instruction- and distribution-invariant preference representations, while routed experts model specialized preferences. This "general + specialized" structure provides a stable preference baseline while retaining expert diversity, improving the bias--variance trade-off in reward modeling.

\subsection{Training the MoE Reward Model}
We initialize the MoE reward model directly from an SFT checkpoint. For each FFN layer that is upcycled, we copy its parameters into both the shared expert and all routed experts, so that the experts within that layer start from identical FFN parameters. We then train the resulting MoE reward model on preference data.

For the $t$-th token, the input to the MoE layer at layer $\ell$  is denoted as $u_t^{(\ell)}$. We employ $K$ experts in each layer, where the shared expert is indexed by $e=0$ and the standard experts are indexed by $e\in\{1,\dots,K-1\}$. The gating weight of the shared expert is fixed by a hyperparameter  $\alpha\in[0,1]$ and is activated for all tokens:
\[
g_{0,t}^{(\ell)}=\alpha.
\]
For the standard experts, the router produces routing logits $r_{e,t}^{(\ell)}$. Let $\mathcal{T}_t^{(\ell)}$ denote the set of top-$m$ experts selected from $\{1,\dots,K-1\}$ (where $m$ is the number of sparsely activated experts), and a softmax normalization is applied within this set:	
\[
\pi_{e,t}^{(\ell)}=
\frac{\exp\!\left(r_{e,t}^{(\ell)}\right)}
{\sum_{j\in \mathcal{T}_t^{(\ell)}} \exp\!\left(r_{j,t}^{(\ell)}\right)},
\quad e\in\mathcal{T}_t^{(\ell)}.
\]
The remaining gating mass $1-\alpha$ is allocated to the experts:
\[
g_{e,t}^{(\ell)}=
\begin{cases}
(1-\alpha)\,\pi_{e,t}^{(\ell)}, & e\in\mathcal{T}_t^{(\ell)},\\
0, & \text{otherwise},
\end{cases}
.
\]
Therefore, the output of this layer is given by
\[
\begin{aligned}
h_t^{(\ell)}
=\alpha\,\mathrm{FFN}_{0}^{(\ell)}\!\left(u_t^{(\ell)}\right)
+\sum_{e=1}^{K-1} g_{e,t}^{(\ell)}\,\mathrm{FFN}_{e}^{(\ell)}\!\left(u_t^{(\ell)}\right),
\\
g_{0,t}^{(\ell)}+\sum_{e=1}^{K-1} g_{e,t}^{(\ell)}=1.
\end{aligned}
\]

\subsection{Parameter Merging from MoE to Dense Models}

After training, to suppress experts that are susceptible to reward hacking, we merge the MoE layer in the parameter space via weighted averaging, thereby recovering a standard feed-forward network (FFN) layer and constructing a dense reward model. To preserve the shared-preference anchor used during MoE training, we fix the merging weight of the shared expert at $\alpha$, while learning the weights of the ordinary experts on a preference calibration set.
Specifically, for layer $\ell$, the raw parameter vector $\boldsymbol{v}^{(\ell)} \in \mathbb{R}^{K-1}$ is normalized via softmax and then multiplied by the weight $1-\alpha$ remaining after allocating $\alpha$ to the shared expert:

\begin{equation}
\bar g_e^{(\ell)}
=(1-\alpha)
\operatorname{softmax}(\mathbf v^{(\ell)})_e,
\quad
\sum_{e=1}^{K-1}\bar g_e^{(\ell)}=1-\alpha .
\end{equation}
Here, $\boldsymbol{v}^{(\ell)}$ is initialized as a zero vector, in which case the softmax output reduces to a uniform distribution, and $\bar{g}_{e}^{(\ell)}$ degenerates to an equal-weight allocation across the $K-1$ ordinary experts. As training proceeds, $\boldsymbol{v}^{(\ell)}$ is updated, and the merging weights deviate from the uniform distribution accordingly, adaptively reflecting the differing contributions of individual experts to the preference discrimination task.

During calibration, only the merging-weight parameters $\boldsymbol{v}^{(\ell)}$ are optimized; all other model parameters remain frozen, and the shared expert's merging weight remains fixed at $\alpha$.

Let $\Theta$ denote the parameters of each expert. Then, the parameters of the merged dense FFN at layer $\ell$ are given by	
\[
\Theta_{\text{merge}}^{(\ell)}
= \alpha\,\Theta_{0}^{(\ell)} + \sum_{e=1}^{K-1}\bar g_{e}^{(\ell)}\,\Theta_{e}^{(\ell)}.
\]

We interpret parameter-space merging as an ensemble-inspired approximation rather than an exact ensemble of expert outputs. Weighted averaging can temper extreme parameter values from individual experts and may improve the robustness of the resulting reward function. When expert-specific errors are sufficiently diverse and compatible with parameter averaging, merging may reduce variance, potentially at the cost of additional bias. This provides a possible mechanism for mitigating brittle reward responses that policies could exploit during PPO. Collapsing the experts into a single dense model also eliminates inference-time sparse routing overhead.

Overall, the proposed merge is designed to promote (i) stability, by retaining a shared-expert parameter anchor with a fixed merging weight $\alpha$; (ii) overfitting suppression, by potentially reducing reliance on expert-specific spurious features; and (iii) reward smoothness, by encouraging less brittle reward responses. These are intended effects rather than guaranteed properties: a fixed parameter-space weight does not imply a fixed functional contribution, and their realization depends on the learned experts and calibration data.
	
\begin{figure*}[t]
\centering

\begin{subfigure}[b]{0.24\linewidth}
    \centering
    \includegraphics[width=\textwidth]{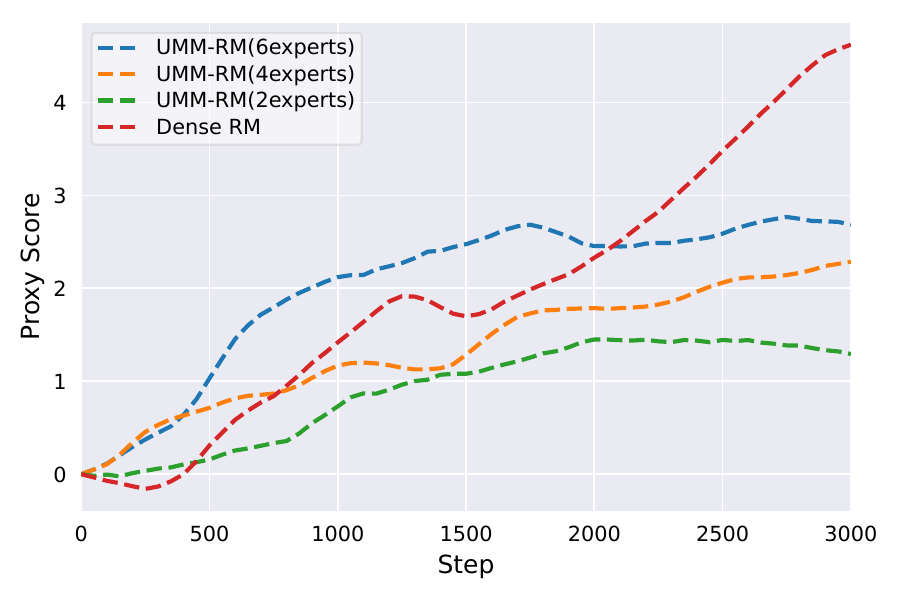}
    \caption{TinyLlama-1.1B}
    \label{fig:fig-umm-t-proxy}
\end{subfigure}
\hfill
\begin{subfigure}[b]{0.24\linewidth}
    \centering
    \includegraphics[width=\textwidth]{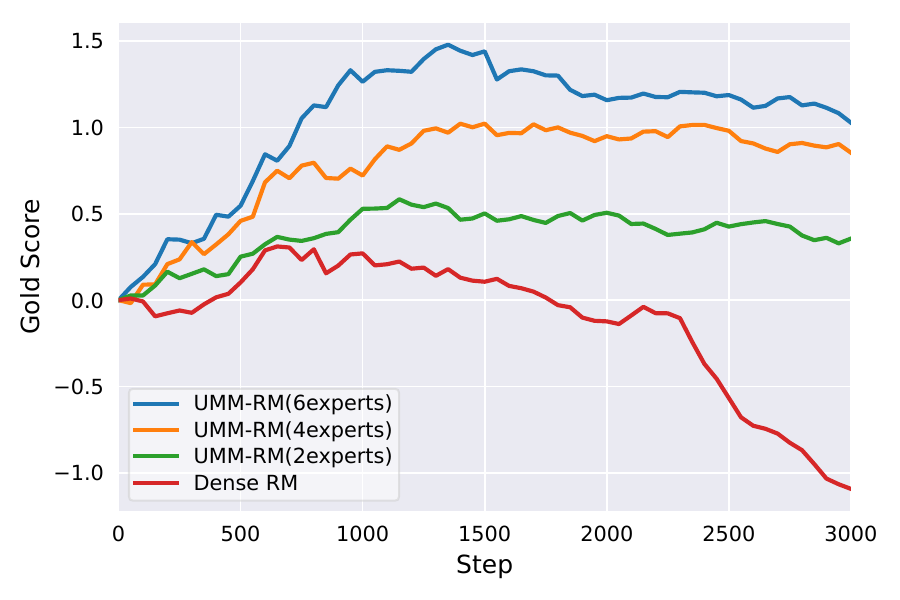}
    \caption{TinyLlama-1.1B}
    \label{fig:fig-umm-t-gold}
\end{subfigure}
\hfill
\begin{subfigure}[b]{0.24\linewidth}
    \centering
    \includegraphics[width=\textwidth]{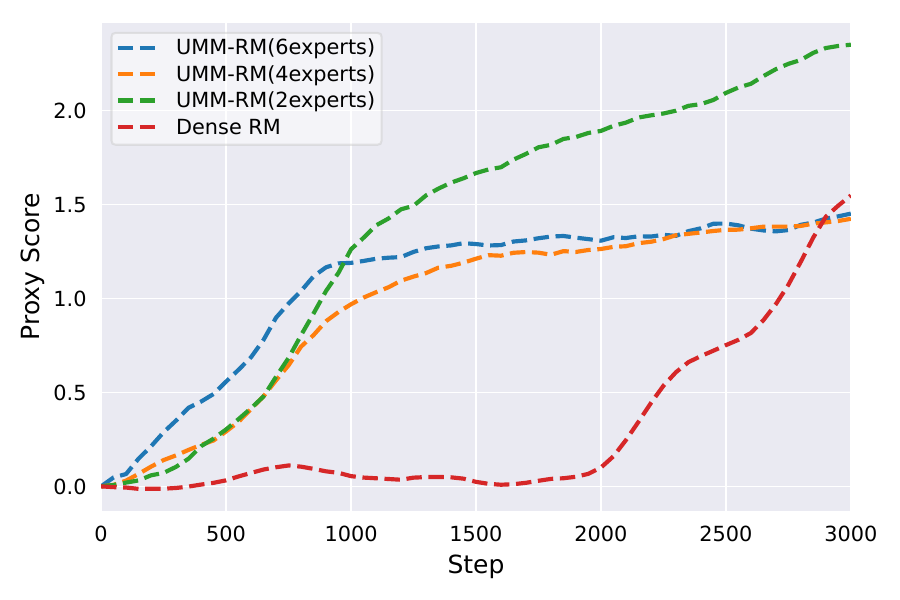}
    \caption{Pythia-1.4B}
    \label{fig:fig-umm-p-proxy}
\end{subfigure}
\hfill
\begin{subfigure}[b]{0.24\linewidth}
    \centering
    \includegraphics[width=\textwidth]{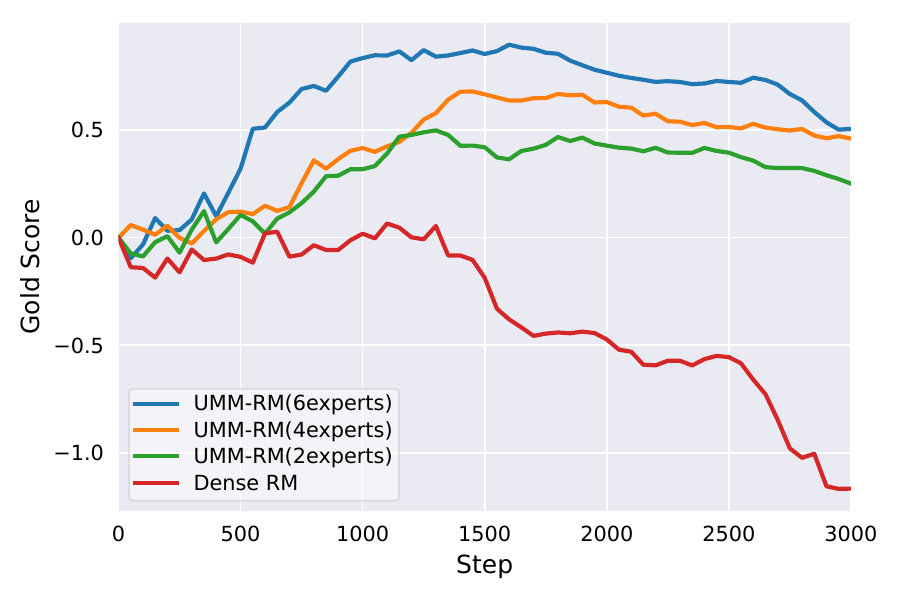}
    \caption{Pythia-1.4B}
    \label{fig:fig-umm-p-gold}
\end{subfigure}

\vspace{0.5em} 

\begin{subfigure}[b]{0.24\linewidth}
    \centering
    \includegraphics[width=\textwidth]{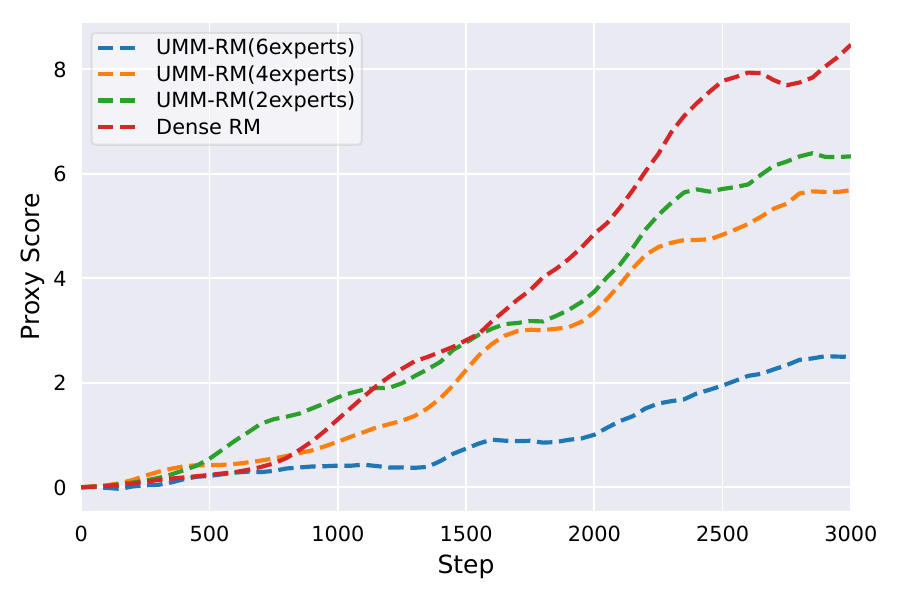}
    \caption{Qwen2.5-0.5B}
    \label{fig:fig-umm-q05-proxy}
\end{subfigure}
\hfill
\begin{subfigure}[b]{0.24\linewidth}
    \centering
    \includegraphics[width=\textwidth]{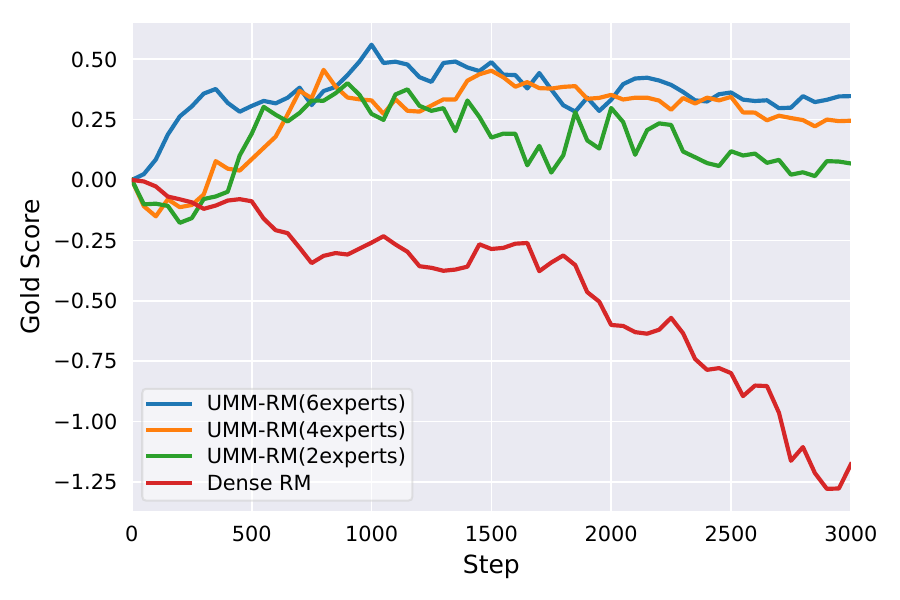}
    \caption{Qwen2.5-0.5B}
    \label{fig:fig-umm-q05-gold}
\end{subfigure}
\hfill
\begin{subfigure}[b]{0.24\linewidth}
    \centering
    \includegraphics[width=\textwidth]{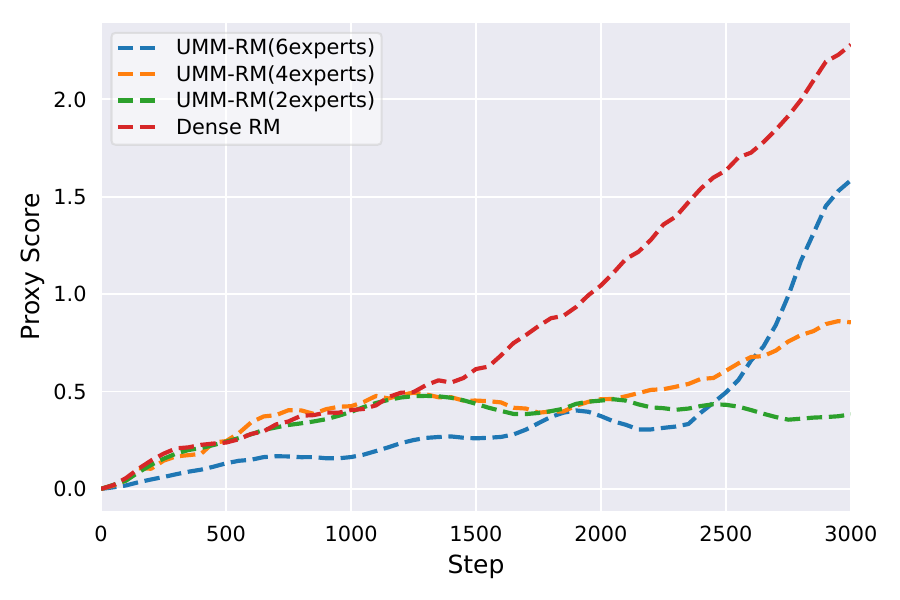}
    \caption{Qwen2.5-1.5B}
    \label{fig:fig-umm-q15-proxy}
\end{subfigure}
\hfill
\begin{subfigure}[b]{0.24\linewidth}
    \centering
    \includegraphics[width=\textwidth]{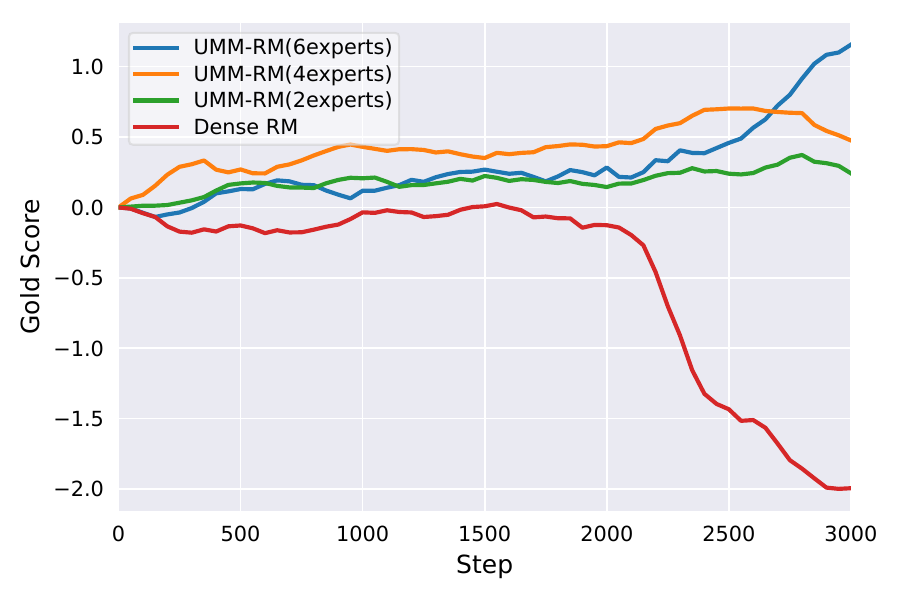}
    \caption{Qwen2.5-1.5B}
    \label{fig:fig-umm-q15-gold}
\end{subfigure}

\caption{Compared with the dense reward model, UMM-RM more effectively mitigates reward hacking. The dashed curves denote the reward scores assigned by the proxy reward model, while the solid curves represent the gold reward scores evaluated by the LLaMA2-13B reward model.}
\label{fig:fig-umm-all}
\end{figure*}

\section{Experiments}
\subsection{Datasets}
We evaluate UMM-RM's ability to mitigate reward hacking on AlpacaFarm \cite{Alpacafarm2023Dubois}, a dataset widely used for RLHF training and evaluation \cite{Reward2023Coste,Inform2024Miao,Finetuning2024Lang}. It contains approximately 52k open-ended and closed-form examples, each with an instruction, optional input, and example output, and provides standard splits and human preference annotations.

We evaluate RM scoring accuracy on Anthropic Helpful and Harmless (HH) \cite{Training2022Bai} and the Webgpt\_comparisons subset of Unified-Feedback \cite{jiang2023llm}. For both HH harmless-base and helpful-base, SFT uses the first 12k training examples (10k train, 2k validation), and RM training uses examples 12,000--22,500 (10k train, 500 validation). Evaluation uses the first 500 examples from each test split.

For Webgpt\_comparisons, SFT uses the first 11k training examples (10k train, 1k validation). RM training uses the next 7k examples, with the remaining training data used for validation. We assess generalization using the official validation split as the test set.
	
\subsection{Models}

To evaluate the proposed method across diverse model scales and architectures, we conduct comparative experiments using four base models: Qwen2.5-0.5B and Qwen2.5-1.5B \cite{qwen2025qwen25technicalreport}, TinyLlama-1.1B \cite{zhang2024tinyllamaopensourcesmalllanguage}, and Pythia-1.4B \cite{biderman2023pythiasuiteanalyzinglarge}. For each backbone, we train an optimized UMM-RM reward model to assess robustness and generality across capacities and designs. We additionally train a larger reward model based on LLaMA2-13B~\cite{touvron2023llama} using the same preference data. We use this model as the Gold RM, whose reward scores serve as a proxy for human judgments during subsequent training and evaluation.

\subsection{Evaluation Metrics}
We report:
(1) preference accuracy on HH and WebGPT;
(2) PPO proxy and gold rewards, with gold rewards provided by a larger LLaMA2-13B-based reward model trained on the same preference data;
(3) GPT-4.1 nano pairwise win/tie/loss rates (temperature=0.3, top-$p$=0.9);
and (4) training/inference efficiency. Inference is measured on the
same GPU in BF16 with batch size 32 and sequence length 520 over 100 runs.

\subsection{RLHF Pipeline}

Our RLHF pipeline broadly follows \cite{Reward2023Coste}, with minor implementation differences. The setup is as follows.

\subsubsection{SFT}
We perform supervised fine-tuning of both the policy and proxy reward models on 10k instruction-following examples from the AlpacaFarm \texttt{sft} split to improve instruction adherence. We use full-parameter fine-tuning for 3 epochs (learning rate $5\times10^{-6}$, batch size 16, max sequence length 512) with Adam ($\beta_1=0.9,\ \beta_2=0.95$) and 600 warm-up steps.

\subsubsection{Reward Model}
Initialized from the SFT checkpoint, the reward model is trained on the AlpacaFarm \texttt{alpaca\_human\_preference} split, holding out 500 samples for evaluation and using the remainder for training. We run full-parameter fine-tuning for 3 epochs (learning rate $3\times10^{-5}$, batch size 32, max sequence length 520) with Adam ($\beta_1=0.9,\ \beta_2=0.95$) and a 0.03 warm-up ratio.

\subsubsection{UMM-RM}
UMM-RM is initialized from the same SFT checkpoint as the dense RM baseline. Each MoE layer contains eight experts: one always-active shared expert and seven routed experts, all initialized by copying the corresponding FFN parameters from the SFT checkpoint. The resulting MoE reward model is then trained on preference data using the same training hyperparameters as the dense RM baseline. We vary the number of routed experts activated per token during MoE training, using $m \in \{2,4,6\}$. All experts are subsequently merged into a single dense RM for PPO reward scoring. Unless otherwise specified, we set the shared-expert weight to $\alpha=0.75$ during both MoE training and parameter merging. For calibration, we sample 1,000 preference pairs from the reward-model training set. During this stage, we optimize only the merging weights of the routed experts, while keeping all other model parameters frozen and the shared expert's merging weight fixed at $\alpha$. We optimize these weights for up to 200 steps with a learning rate of $10^{-4}$ and no warm-up.

\subsubsection{PPO}
We train and evaluate PPO on 20k samples from the AlpacaFarm \texttt{unlabeled} split (\texttt{alpaca\_instructions} subset), reserving 2k for evaluation. Each iteration samples 256 rollouts (chunk size 32, temperature 1) and applies four gradient updates (batch size 32). Training runs for 3{,}000 steps with clipping range 0.2, and GAE $\lambda=0.95$. Following the experimental findings of \cite{Reward2023Coste}, we adopt a KL coefficient of 0.01. Optimization uses AdamW (learning rate $2\times10^{-6}$, $\beta_1=0.9,\ \beta_2=0.95$, weight decay $10^{-6}$) with cosine annealing. The policy and value functions share a backbone and branch into separate heads only after the final two layers to output the action distribution and value estimates.

\subsection{Experimental Results}
\subsubsection{Evaluation on RLHF}

We evaluate three UMM-RM variants trained with $m=2,4,6$ activated routed experts per token. Throughout, the labels ``2-experts'', ``4-experts'', and ``6-experts'' refer to this MoE training configuration, excluding the always-active shared expert; the merged reward models used during PPO are dense. Experiments span multiple base models and compare UMM-RM against a dense RM and ensemble RM baselines using Mean/WCO/UWO \cite{Reward2023Coste} and WARM \cite{Warm2024Rame}. For consistency, all methods’ reward scores are shifted so the initial starts at zero.

Figure~\ref{fig:fig-umm-all} summarizes the performance of UMM-RM as an RLHF reward model. Using TinyLlama-1.1B as an example (Figs.~\ref{fig:fig-umm-t-proxy} and~\ref{fig:fig-umm-t-gold}), the dense RM’s proxy score exhibits an overall upward trend during training, whereas its gold score rises early and then collapses, reflecting classic reward hacking. By contrast, UMM-RM achieves higher gold scores that decline after reaching their peaks but remain well above those of the dense RM at the end of training. These results suggest that UMM-RM largely mitigates reward hacking, though minor residual effects remain. In addition, enabling more experts (4/6) suppresses reward hacking more effectively than enabling fewer (2).

\begin{figure}[t]
    \centering
    \begin{subfigure}[b]{0.48\linewidth}
        \centering
        \includegraphics[width=\textwidth]{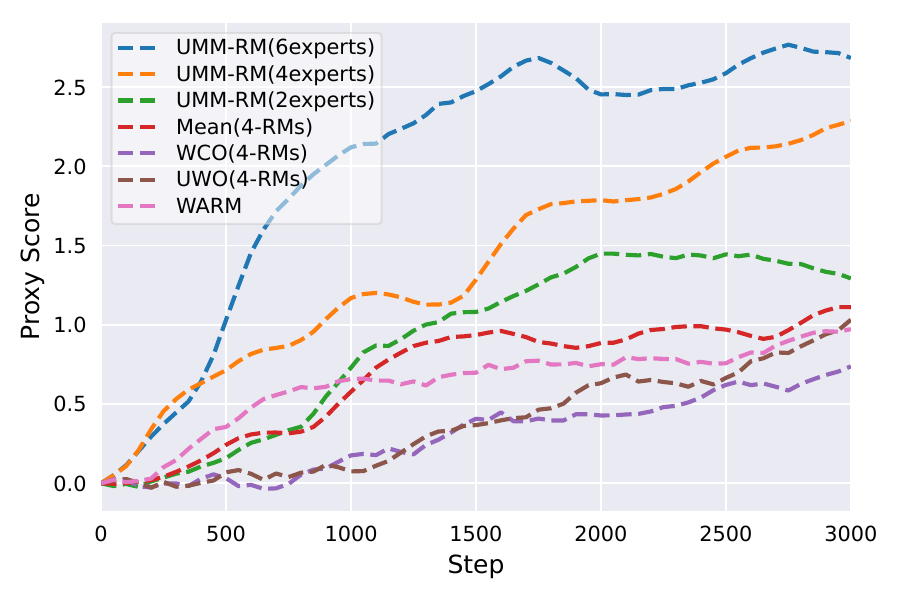}
        \caption{TinyLlama-1.1B}
        \label{fig:fig-umm-t-other-proxy}
    \end{subfigure}
    \hfill
    \begin{subfigure}[b]{0.48\linewidth}
        \centering
        \includegraphics[width=\textwidth]{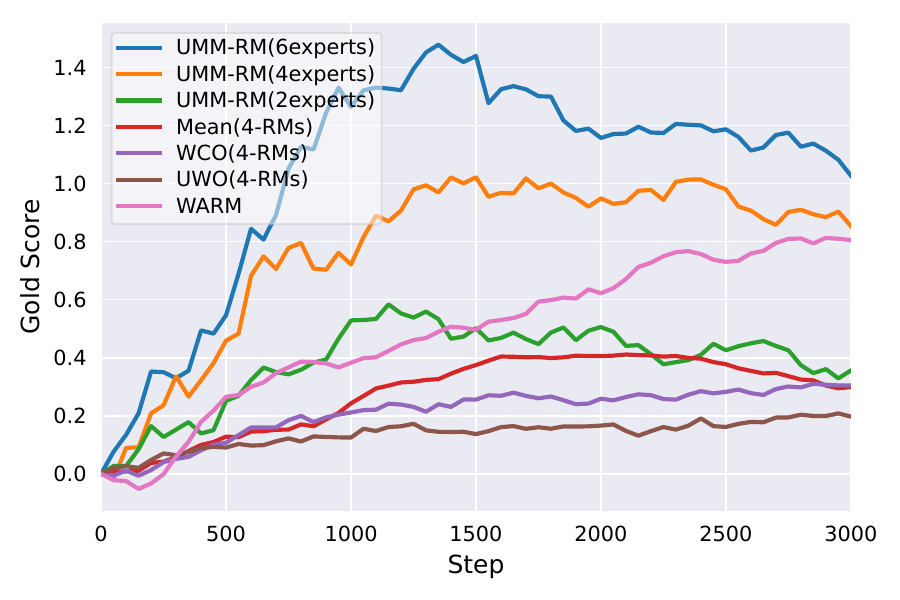}
        \caption{TinyLlama-1.1B}
        \label{fig:fig-umm-t-other-gold}
    \end{subfigure}
    \caption{Comparison of PPO training reward scores between UMM-RM and reward models based on different ensemble strategies built on the TinyLlama-1.1B backbone.}
    \label{fig:fig-umm-t-other}
\end{figure}

On TinyLlama-1.1B, we further compare UMM-RM with ensemble RM methods \cite{Reward2023Coste} and WARM \cite{Warm2024Rame}. The ensemble RM trains four reward models with different random seeds and aggregates them using Mean, WCO, and UWO, respectively; WARM ensembles three reward models trained with different learning rates. As shown in Fig.~\ref{fig:fig-umm-t-other}, UMM-RM with 4 or 6 activated routed experts generally achieves higher gold scores than the Mean/WCO/UWO ensemble RMs and WARM during PPO training. The 2-expert variant also performs favorably against Mean/WCO/UWO overall, although its gold-score curve crosses that of Mean at some training steps. These results demonstrate that, by introducing structured expert diversity within a single reward model, UMM-RM provides a more stable reward signal and effectively suppresses reward hacking while maintaining inference efficiency.

\subsubsection{Accuracy Evaluation}

We evaluate TinyLlama-1.1B–based UMM-RM against Ensemble RM \cite{Reward2023Coste} on Anthropic HH and the Webgpt\_comparisons subset of Unified-Feedback (Table~\ref{tab:table0}). With 6 activated experts, UMM-RM achieves 58.4\%, 57.6\%, and 60.8\% accuracy on HH-Harmless, HH-Helpful, and Webgpt\_comparisons, respectively, exceeding the Mean, WCO, and UWO ensemble results.

Further experiments on Qwen2.5-0.5B and Pythia-1.4B (Table~\ref{tab:table1}) show that all UMM-RM configurations outperform the dense RM across all three backbones and datasets. Qwen2.5-0.5B achieves the highest HH-Helpful accuracy of 67.2\%, while Pythia-1.4B with 6 activated experts reaches 55.2\%, 54.8\%, and 57.8\% on the three datasets, respectively. These results demonstrate consistent accuracy gains across the evaluated backbones and preference datasets.

\begin{table}[t]
\centering
\small
\setlength{\tabcolsep}{2.5pt}
\begin{tabular}{lccc}
\toprule
Models & \multicolumn{2}{c}{Anthropic} & WebGPT \\
\cmidrule(lr){2-3}
& Harmless & Helpful &  \\
\midrule
 Dense RM             & 51.2$\pm$0.3 & 44.6$\pm$0.2 & 52.2$\pm$0.7 \\
 
\makecell[l]{Mean (ensemble RM)}    & 57.1$\pm$0.3 & 55.0$\pm$0.3 & 51.4$\pm$0.2 \\
\makecell[l]{WCO (ensemble RM)}   & 55.4$\pm$0.6 & 54.8$\pm$0.2 & 60.6$\pm$0.3 \\
\makecell[l]{UWO (ensemble RM)}  & 58.0$\pm$0.4 & 54.6$\pm$0.5 & 59.6$\pm$0.4 \\

\midrule
UnMerged MoE (2-experts) & \textbf{59.8$\pm$0.2} & 53.8$\pm$0.2 & 57.0$\pm$0.3 \\
UnMerged MoE (4-experts) & 58.2$\pm$0.6 & 55.8$\pm$0.2 & 58.2$\pm$0.5 \\
UnMerged MoE (6-experts) & 57.4$\pm$0.4 & 56.2$\pm$0.2 & 60.2$\pm$0.2 \\

\midrule
UMM-RM (2-experts)          & 56.4$\pm$0.4 & 54.2$\pm$0.2 & 57.8$\pm$0.5 \\
UMM-RM (4-experts)          & 58.0$\pm$0.3 & 55.2$\pm$0.3 & 58.6$\pm$0.2 \\
UMM-RM (6-experts)          & 58.4$\pm$0.4 & \textbf{57.6$\pm$0.2} & \textbf{60.8$\pm$0.4} \\
\bottomrule
\end{tabular}

\caption{Preference accuracy (\%) of different reward modeling methods on TinyLlama-1.1B. Results are reported as mean $\pm$ std. over 3 random seeds.}
\label{tab:table0}
\end{table}

\begin{table}[htbp]
    \centering
    \small
    \setlength{\tabcolsep}{3.5pt}
    \begin{tabular}{lccc}
        \toprule
        Models & \multicolumn{2}{c}{Anthropic} & WebGPT \\
        \cmidrule(lr){2-3}
        & Harmless & Helpful &  \\
        \midrule
        \multicolumn{4}{c}{Base Model: Qwen2.5-0.5B} \\
        \midrule
        Dense RM & 38.6 & 65.8 & 57.2 \\
        UMM-RM (2-experts) & 48.8 & 66.8 & 58.2 \\
        UMM-RM (4-experts) & 50.2 & 66.0 & 57.8 \\
        UMM-RM (6-experts) & \textbf{50.8} & \textbf{67.2} & \textbf{58.4} \\
        \midrule
        \multicolumn{4}{c}{Base Model: Pythia-1.4B} \\
        \midrule
        Dense RM & 48.0 & 44.6 & 50.8 \\
        UMM-RM (2-experts) & 54.6 & 53.0 & 54.2 \\
        UMM-RM (4-experts) & 54.2 & 53.4 & 54.0 \\
        UMM-RM (6-experts) & \textbf{55.2} & \textbf{54.8} & \textbf{57.8} \\
        \bottomrule
    \end{tabular}
    \caption{Comparison of accuracy (\%) on preference tasks across different reward models using Qwen2.5-0.5B and Pythia-1.4B.}
    \label{tab:table1}
\end{table}

\subsubsection{Unmerged MoE vs. UMM-RM}

\begin{figure}[th]
    \centering
    \begin{subfigure}[b]{0.48\linewidth}
        \centering
        \includegraphics[width=\textwidth]{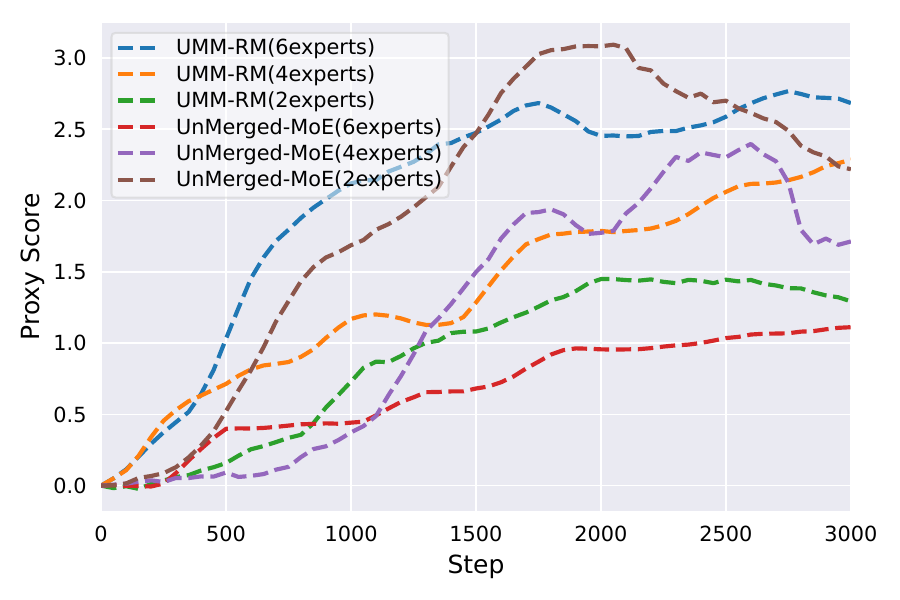}
        \caption{TinyLlama-1.1B}
        \label{fig:fig-umm-t-umerge-proxy}
    \end{subfigure}
    \hfill
    \begin{subfigure}[b]{0.48\linewidth}
        \centering
        \includegraphics[width=\textwidth]{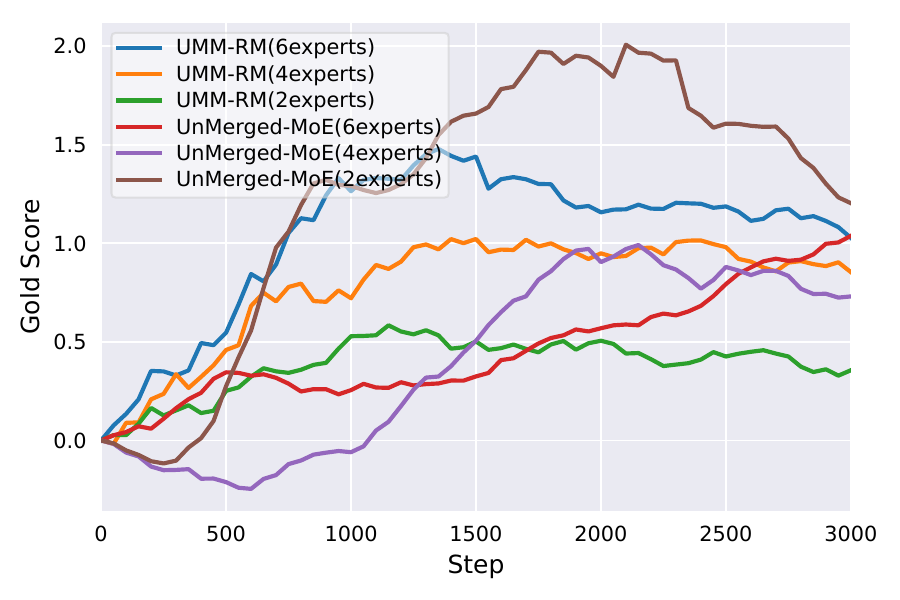}
        \caption{TinyLlama-1.1B}
        \label{fig:fig-umm-t-umerge-gold}
    \end{subfigure}
    \caption{Comparison of PPO training reward scores between MoE reward models with different numbers of activated experts and UMM-RM based on the TinyLlama-1.1B backbone.}
    \label{fig:fig-umm-t-umerge}
\end{figure}

Table~\ref{tab:table0} and Fig.~\ref{fig:fig-umm-t-umerge} compare UMM-RM with unmerged MoE in preference accuracy and PPO robustness. The 2-expert unmerged MoE achieves the best HH-Harmless accuracy (59.8\%), suggesting that direct routing can preserve useful specialization in low-expert settings. However, with six activated experts, UMM-RM outperforms the corresponding unmerged MoE on all three benchmarks. The PPO results show a more configuration-dependent effect. In the 4-expert setting, UMM-RM reaches higher gold rewards earlier and maintains a relatively stable plateau. In the 6-expert setting, UMM-RM reaches a higher peak gold reward but subsequently declines, whereas the unmerged MoE's gold reward generally increases throughout training.

\subsubsection{Effect of the Shared-Expert Weight}

\begin{figure}[th]
\centering
\begin{subfigure}[b]{0.48\linewidth}
    \centering
    \includegraphics[width=\textwidth]{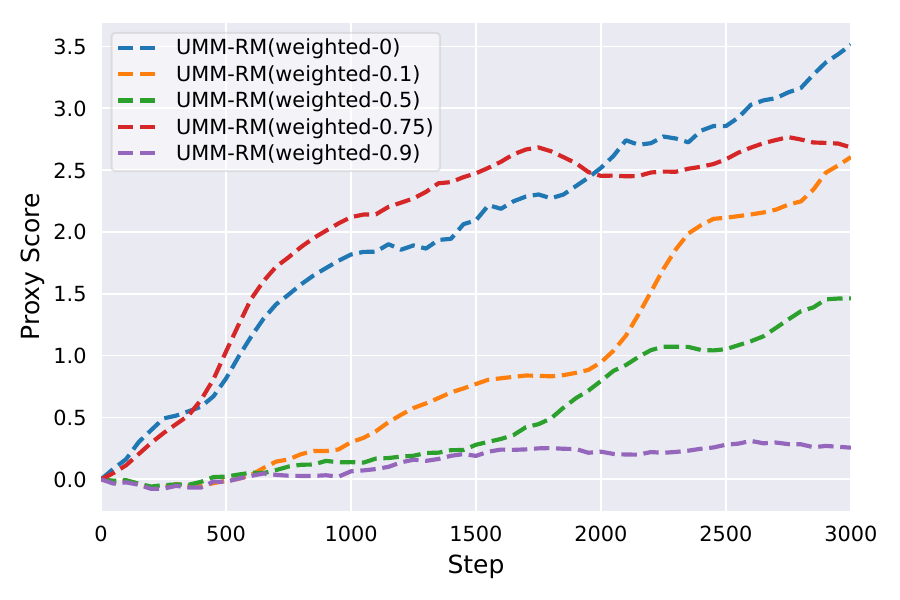}
    \caption{TinyLlama-1.1B}
    \label{fig:proxy}
\end{subfigure}
\hfill
\begin{subfigure}[b]{0.48\linewidth}
    \centering
    \includegraphics[width=\textwidth]{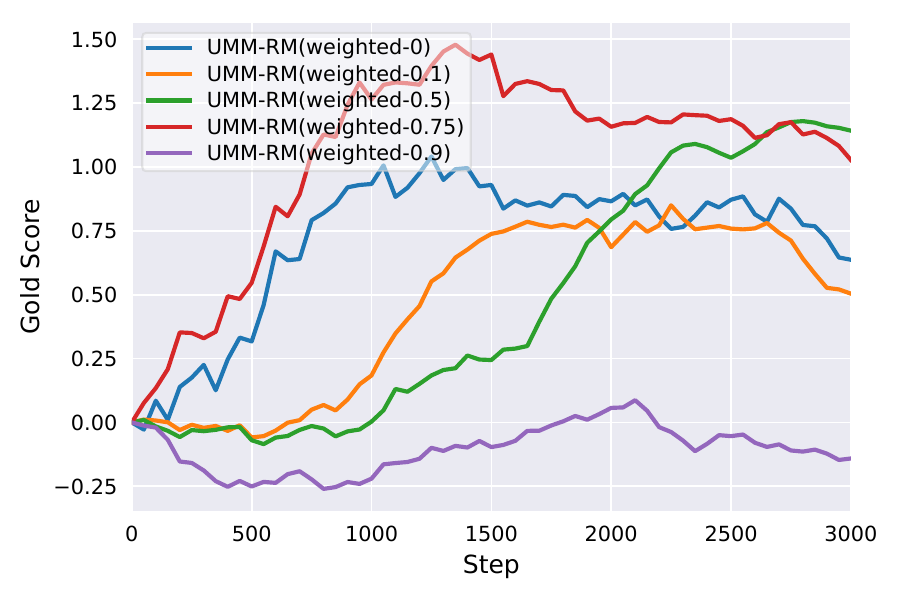}
    \caption{TinyLlama-1.1B}
    \label{fig:gold}
\end{subfigure}
\caption{Proxy (left) and gold (right) reward scores during PPO training for UMM-RM with a TinyLlama-1.1B backbone and shared-expert weights $\alpha \in \{0.1, 0.5, 0.75, 0.9\}$.}
\label{fig:t-expert-weighted}
\end{figure}

Using TinyLlama-1.1B as the backbone, we study the effect of the shared-expert weight $\alpha$ in the UMM-RM setup with six activated routed experts during PPO training (Fig.~\ref{fig:t-expert-weighted}). We compare the four settings shown in the figure: $\alpha \in \{0.1, 0.5, 0.75, 0.9\}$. Among these settings, $\alpha=0.5$ achieves the highest gold score at the final plotted training step (step 3000), whereas $\alpha=0.75$ reaches the highest gold score over the full training trajectory but subsequently declines. The $\alpha=0.9$ setting yields the lowest final gold score. For $\alpha=0.1$, the proxy score increases substantially during later training while the gold score declines, indicating reward hacking. These results highlight the sensitivity of PPO performance to the shared-expert weight.

\begin{table}[t]
    \centering
    \small
    \setlength{\tabcolsep}{2pt}
    \begin{tabular}{@{}llccc@{}}
        \toprule
        \multirow{2}{*}{Models} & \multirow{2}{*}{Opponent}
        & \multicolumn{3}{c}{AlpacaFarm} \\
        \cmidrule(lr){3-5}
        & & Win$\uparrow$ & Tie & Lose$\downarrow$ \\
        \midrule

        UMM-RM(2-experts) & \multirow{3}{*}{Dense RM}
        & 28.9 & 44.0 & 27.1 \\
        UMM-RM(4-experts) &
        & 30.6 & 45.5 & 23.9 \\
        UMM-RM(6-experts) &
        & 36.4 & 42.3 & 21.3 \\

        \midrule
        UMM-RM(2-experts) & \multirow{3}{*}{SFT Model}
        & 32.3 & 39.9 & 27.8 \\
        UMM-RM(4-experts) &
        & 33.7 & 37.5 & 28.8 \\
        UMM-RM(6-experts) &
        & 36.4 & 34.3 & 29.3 \\

        \midrule
        \multirow{3}{*}{UMM-RM(4-experts)}
        & Mean & 11.8 & 71.1 & 17.1 \\
        & WCO  & 27.0 & 63.4 & 9.6 \\
        & UWO  & 16.5 & 73.9 & 9.6 \\

        \midrule
        \multirow{3}{*}{UMM-RM(6-experts)}
        & Mean & 22.0 & 60.4 & 17.6 \\
        & WCO & 28.3 & 60.2 & 11.5 \\
        & UWO  & 23.9 & 62.4 & 13.7 \\

        \bottomrule
    \end{tabular}
    \caption{End-to-end generation quality evaluation of policy models trained with PPO under different reward model methods.}
    \label{tab:table3}
\end{table}
\subsubsection{End-to-End Generation Quality Evaluation}

We evaluate how different reward models mitigate reward hacking using the \texttt{alpaca\_human\_evaluation} split of the AlpacaFarm dataset, with TinyLlama-1.1B as the base model. We compare PPO-trained policies using UMM-RM, a standard dense RM, and ensemble RMs (Mean, WCO, UWO), along with the corresponding SFT baseline. Results are shown in Table~\ref{tab:table3}.

Overall, UMM-RM yields higher win rates than both the dense RM and SFT, suggesting it steers PPO toward responses aligned with human preferences rather than simply maximizing reward scores. Increasing the number of routed experts activated during MoE training yields merged reward models whose PPO-trained policies achieve higher win rates against the dense-RM-trained policy and the SFT baseline in these experiments. Notably, the 6-expert UMM-RM outperforms the Mean, WCO, and UWO ensembles, implying that the upcycle-and-merge shared-expert design provides more effective reward supervision than simple ensembling. While the 4-expert UMM-RM slightly trails the Mean ensemble, the six-activated-expert training configuration yields a merged RM whose PPO-trained policy achieves higher pairwise win rates than the four-activated-expert configuration against all three ensemble baselines.

\subsubsection{Training and Inference Efficiency}
Table~\ref{tab:efficiency} compares the efficiency of different
reward modeling methods on TinyLlama-1.1B. Although UMM-RM requires
about $2\times$ the training time of the dense RM, its merged model
achieves nearly identical inference latency and throughput
(381.9 ms/batch and 83.78 samples/s). By contrast, the unmerged MoE
and four-model ensemble incur $2.39\times$ and $4.04\times$ inference
latency, respectively. This demonstrates that UMM-RM trades moderate
training overhead for dense-level deployment efficiency.

\begin{table}[t]
\centering
\small
\setlength{\tabcolsep}{1.5pt}
\begin{tabular}{lcccc}
\toprule
Method &
Train &
Latency &
Throughput &
Slowdown \\
\midrule

Dense RM
& $\sim$1.0 h& 382.2$\pm$1.1& 83.72& $1.00\times$ \\

Unmerged MoE
& $\sim$1.5 h& 914.4$\pm$4.3& 35.00& $2.39\times$ \\

Ensemble RM (4)
& $\sim$4.0 h& 1543.5$\pm$3.7& 20.73& $4.04\times$ \\

UMM-RM
& $\sim$2.0 h& 381.9$\pm$1.1& 83.78& $1.00\times$ \\
\bottomrule
\end{tabular}
\caption{Training and inference efficiency comparison of different reward modeling methods based on TinyLlama-1.1B. Slowdown is measured relative to Dense RM.}
\label{tab:efficiency}
\end{table}

\section{Related Work}

\subsection{Mitigating Reward Hacking in RLHF}

Reward hacking arises when the learned reward model deviates from true human preferences, allowing the policy to exploit such discrepancies under closed-loop optimization, especially when reward models are under-parameterized or trained on limited data~\cite{Scaling2023Gao}. Existing mitigation methods mainly improve reward-model robustness, use ensembles or uncertainty-aware objectives, or introduce algorithmic safeguards against over-optimization.

One line of work strengthens individual reward models. \cite{Towards2024Qin} use confidence-aware training to reduce errors from ambiguous or noisy labels, while \cite{yang2024regularizing} improve out-of-distribution generalization through hidden-state regularization and an auxiliary text-generation loss. PIRA~\cite{xyf2025pira} formulates reward modeling as instruction-guided preference evaluation with explicit criteria and reduces reward variance through averaging across evaluation instructions and stochastic value-head outputs. However, single-model approaches can still inherit systematic reward biases.

Another line of work mitigates over-optimization through reward-model ensembles. \cite{Reward2023Coste} study mean optimization (Mean), worst-case optimization (WCO), and uncertainty-weighted objectives (UWO). \cite{Scalable2024Ahmed} reduce ensemble cost using a shared backbone with multiple linear heads, while \cite{ImprovingZhang} compare linear and LoRA-based ensembles and find LoRA ensembles more effective against reward hacking. \cite{Warm2024Rame} improve OOD reliability by averaging reward-model parameters trained under different conditions, whereas \cite{Helping2023Eisenstein} emphasize the importance of diversity in reward signals. Despite improved robustness, ensemble approaches often introduce additional training or inference costs.

More recent work explores MoE-based reward models to combine specialization and diversity. \cite{Dmoerm2024Quan} proposes a two-level MoE reward model with sparse routing and MLP-based expert fusion, while \cite{Interpretable2024Wang} gate experts using prompt semantics and multi-dimensional evaluations. These methods enhance preference modeling capacity, but sparse routing and architectural complexity generally increase inference overhead.

\subsection{Upcycled MoE Model}

Sparse Mixture-of-Experts models increase model capacity without proportionally increasing computation, but training them from scratch remains costly. Upcycling methods therefore initialize sparse MoE models from dense checkpoints. \cite{Sparse2022Komatsuzaki} show that sparse upcycling can substantially reduce pretraining cost. \cite{XFT2024Ding} further show that naive sparse upcycling provides limited gains for instruction tuning and propose XFT, which introduces shared experts and routing-weight normalization during MoE training before merging experts back into a dense model. Similarly, \cite{One2022Xue} demonstrate that knowledge from sparse experts can be transferred into dense models.

UMM-RM follows this general upcycle-and-merge paradigm but adapts it specifically to reward modeling. It maintains a fixed shared-expert contribution during MoE training and performs pairwise-preference-calibrated expert merging, with the goal of reducing exploitable reward responses during PPO while retaining dense-model inference efficiency.

\section{Conclusions}
In this study, we propose UMM-RM, a Mixture-of-Experts (MoE) reward modeling approach based on an upcycle-and-merge strategy, to mitigate reward hacking during RLHF training. UMM-RM initializes MoE experts by copying FFN parameters from an SFT checkpoint, trains them on preference data, and merges them into a dense reward model with a parameter count comparable to the original. This design enhances reward modeling capacity while retaining dense-model inference efficiency. Experiments across diverse settings demonstrate that UMM-RM effectively mitigates reward hacking, offering a practical and efficient approach to building reliable reward models.

\section*{Limitations}

This work has two main limitations. First, our experiments are limited
to relatively small and medium-scale backbones, and further validation
on larger language models is needed to confirm the scalability of
UMM-RM. Second, although UMM-RM preserves dense-model inference
efficiency after merging, it introduces extra training overhead during
MoE upcycling and merging-weight optimization.


\end{document}